%% file: main.tex
\documentclass[10pt,twocolumn,letterpaper]{article}

\usepackage{iccv}      
\usepackage[accsupp]{axessibility}  
\usepackage{tabularx}
\usepackage{multicol}
\usepackage{multirow}
\usepackage{xltabular}
\usepackage{algorithm}
\usepackage{algpseudocode}
\usepackage{amsmath}
\usepackage{fancyhdr}
\input{preamble}

\definecolor{iccvblue}{rgb}{0.21,0.49,0.74}
\usepackage[pagebackref,breaklinks,colorlinks,allcolors=iccvblue]{hyperref}

\def\paperID{3} 
\def\confName{ICCV}
\def\confYear{2025}

\title{Do VLMs Have Bad Eyes? Diagnosing Compositional Failures via Mechanistic Interpretability}

\author{%
  Ashwath Vaithinathan Aravindan, Abha Jha, Mihir Kulkarni  \\
  University of Southern California\\
  \texttt{\{vaithina, abhajha, mkulkarn\}@usc.edu} \\
}

\begin{document}
\begin{table*}[h]
\begin{tabular}{m{\textwidth}}
© © 2025 IEEE. Personal use of this material is permitted. Permission from IEEE must be obtained for all other uses, in any current or future media, including reprinting/republishing this material for advertising or promotional purposes, creating new collective works, for resale or redistribution to servers or lists, or reuse of any copyrighted component of this work in other works.\\
\end{tabular}
\end{table*}

\maketitle
\input{sec/0_abstract}    
\input{sec/1_intro}
\input{sec/2_related_works}
\input{sec/3_methods}
\input{sec/4_experiments}
\input{sec/5_results}
\input{sec/6_conclusion}

\input{sec/figures_and_tables}
\clearpage
{
    \small
    \bibliographystyle{ieeenat_fullname}
    \bibliography{references}
}


\end{document}

%% file: preamble.tex
\usepackage{algorithmicx}
\usepackage{algorithm}

%% file: sec/0_abstract.tex
\begin{abstract}Vision-Language Models (VLMs) have shown remarkable performance in integrating visual and textual information for tasks such as image captioning and visual question answering. However, these models struggle with compositional generalization and object binding, which limit their ability to handle novel combinations of objects and their attributes. Our work explores the root causes of these failures using mechanistic interpretability techniques. We show evidence that individual neurons in the MLP layers of CLIP's vision encoder represent multiple features, and this "superposition" directly hinders its compositional feature representation which consequently affects compositional reasoning and object binding capabilities. We hope this study will serve as an initial step toward uncovering the mechanistic roots of compositional failures in VLMs. The code and supporting results can be found \href{https://github.com/Mystic-Slice/Do-VLMs-Have-Bad-Eyes}{here}.
\end{abstract}


%% file: sec/1_intro.tex
\section{Introduction}
\label{sec:intro}

Vision-Language Models (VLMs) have demonstrated impressive capabilities in tasks like image captioning, visual question answering, and zero-shot classification by integrating visual and textual information. However, they often struggle with \textit{compositionality}—the ability to understand and reason about novel combinations of familiar concepts~\cite{compositionality_defn}. In this context, compositionality refers to recognizing combinations of known objects and attributes.

A related issue is the \textit{binding problem}~\cite{binding_problem}, where the model fails to associate specific attributes (e.g., color, shape, position) with the correct object. Robust binding is essential for fine-grained and coherent multimodal understanding.

While techniques have been proposed to encourage Convolutional Neural Networks (CNNs) to develop compositional representations \cite{teach_compo_cnns}, VLMs face additional complexity due to their need to handle both visual and textual data simultaneously. 




We hypothesize that compositional failures in Vision-Language Models (VLMs) stem from \textbf{superposition} \cite{mech_interp_problems}, where multiple concepts are entangled within shared representational subspaces. This entanglement may lead to the incorrect binding of objects, attributes, and relationships within multimodal tasks. To investigate this, we employ mechanistic interpretability techniques to analyze internal representations and diagnose failure points. Our central research question is: \textit{Are entangled or misaligned representations within the CLIP vision encoder the root cause for failures in compositional reasoning?}

We perform several experiments with CLIP's vision encoder and note our observations about feature entanglement at the neuron level. We use Grad-CAM to analyze the spatial attention patterns of CLIP in response to compositional text prompts. These visualizations revealed consistent attribute–object binding failures, where the model incorrectly attended to multiple unrelated regions or failed to isolate the correct object-attribute pair. 


To further investigate the underlying cause of these failures, we conduct a neuron-level analysis of CLIP’s MLP activations. Using a synthetic dataset of simple shapes, colors, and positions, we identified "feature neurons", neurons that respond selectively to specific visual attributes, and used Shannon entropy to quantify their selectivity. 

The results of our experiments strongly suggest that individual neurons encode multiple visual concepts, proving the presence of superposition. We also show that superposition of features in neurons is related to how separable the features are in the output embedding space.



%% file: sec/2_related_works.tex
\section{Related Work}

\textbf{Compositionality and Binding in VLMs.}  
Previous studies~\cite{compositional_challenges, binding_problem} have shown that VLMs struggle with associating attributes like color, shape, and count to specific objects, often leading to compositional failures. While these works document the failures, our work goes further by investigating internal representations—specifically superposition and attention misalignment—as potential causes.\\
\textbf{CLIP Interpretability and Concept Alignment.}  
Recent work~\cite{quant_interp_clip, textspan} quantifies how CLIP-like models align with human-understandable concepts using attention and text decomposition. In contrast, our work is focused on uncovering why these models fail at binding—by tracing failure cases to specific neurons and spatial attention behaviors. We use techniques from tools like Prisma~\cite{prisma_LW, prisma_code} for analyzing the CLIP models.\\
\textbf{Superposition.}
Elhage et al.\cite{elhage2022toy} explores polysemanticity of neurons using small ReLu networks and by limiting the data to sparse input features. Olah et al.\cite{olah2020zoom} expands superposition to vision tasks and proposes that superposition exists across circuits within the network.
Our work focuses on neuron-level superposition and also studies how it affects Object binding and compositionality problems for vision and multimodal tasks. 

%% file: sec/3_methods.tex
\section{Methodology}  
\label{sec:methods}  

\subsection{Datasets}
Following are the datasets we use throughout our experiments:
\begin{enumerate}
    \item CIFAR-10 dataset that contains 60000 images that belong to 10 different classes. \cite{cifar10}
    \item A toy Shapes dataset we created, containing 500 images with simple shapes of different colors.
\end{enumerate}

\subsection{Model}
We use the pretrained \texttt{CLIP-ViT-L/14} ~\cite{clip-vit-large-patch14} model and its associated pre-processor due to the wide use of CLIP and its variants as the vision encoder in most VLMs. We pick the simplest variant to allow easier examination of internal representation. The vision encoder component takes a single RGB image of size 224x224 and outputs an embedding of length 768.

\subsection{Visual Grounding with CLIP and Gradient-Based Localization} \label{subsec:gradcam_visualisation}

Although CLIP \cite{clip-vit-large-patch14} has a strong performance on tasks like image retrieval and zero-shot classification, it often struggles to bind attributes to objects (e.g., confusing 'a red square' with 'a red circle') or correctly interpret spatial and relational signals. One possible reason for these failures is misalignment in the model’s attention mechanisms. That is, even if CLIP correctly identifies relevant features, it may not bind them to the correct object instance. To investigate this, we use gradient-based attribution—specifically Grad-CAM~\cite{gradcam}—to visualize which parts of an image CLIP attends to when making a prediction for a given text prompt.



\subsubsection{Activation and Gradient Extraction}
To understand which visual features influence CLIP’s predictions, we register hooks on the final MLP layer of the vision encoder to capture both activations and gradients. During the forward pass, we record the activation maps for the image. Then, we perform a backward pass on the probability corresponding to a specific text prompt, which gives us the gradient of the prediction with respect to the image features.

\subsubsection{Grad-CAM Computation}

We compute a Grad-CAM heatmap for each prompt using the following steps:

\begin{enumerate}
    \item A forward pass computes the similarity score $S_t$ between the image and a text prompt $t$.
    
    \item A backward pass computes the gradient of $S_t$ with respect to the activation map $\mathbf{A} \in \mathbb{R}^{H \times W \times C}$ from a chosen layer in the vision transformer:
    \[
    \frac{\partial S_t}{\partial \mathbf{A}} \in \mathbb{R}^{H \times W \times C}
    \]
    
    \item We average the gradient across spatial locations to compute importance weights for each channel:
    \[
    \alpha_c = \frac{1}{H \times W} \sum_{i=1}^H \sum_{j=1}^W \frac{\partial S_t}{\partial \mathbf{A}_{ij}^c}
    \]
    
    \item These weights are used to compute a weighted combination of the channels in the activation map:
    \[
    L_{\text{Grad-CAM}} = \text{ReLU} \left( \sum_c \alpha_c \mathbf{A}^c \right)
    \]
\end{enumerate}


This procedure reveals the spatial regions that contribute the most to the CLIP decision for a given text prompt. By analyzing these attention patterns across different prompts and images, we gain insight into how CLIP binds - or fails to bind - semantic attributes to objects in its visual representation.

\subsection{Analyzing CLIP Activations} \label{subsec:clip_act_analysis}

To better understand the behavior of the CLIP vision encoder and to search for evidence of superposition, we focus our analysis on the activations of the MLP layers. MLPs are a natural target for such investigation, as they constitute the majority of a transformer's parameters. Prior work has shown that MLPs in large language models function as key–value stores~\cite{mlp-kv-stores, rome-mlp-kv-stores}, and are responsible for encoding factual and compositional knowledge. We hypothesize that MLPs in the CLIP vision encoder may play a similarly important role in storing visual concepts and their associations.

\subsubsection{Toy Shapes Dataset}
To facilitate this analysis, we construct a toy dataset of 500 synthetic images. Each image contains 1 to 5 objects, with features randomly sampled from a controlled set of attributes: 5 shapes (circle, triangle, square, pentagon, hexagon), 6 colors (red, green, blue, pink, black, yellow), and 5 spatial locations (top-left, top-right, bottom-left, bottom-right, center). A few example images from this dataset are shown in Figure~\ref{fig:toy-dataset-sample}. 

The low visual and semantic complexity of this dataset enables more direct correlation between object-level features and neuron activations in CLIP. This setting allows us to isolate and interpret the behavior of individual neurons, making it easier to identify feature selectivity and potential cases of superposition.

\subsubsection{Methodology}
For our analysis, we record the activation of every neuron in the final linear layer of each MLP block within the CLIP vision encoder. The CLIP-ViT-L/14 model~\cite{clip-vit-large-patch14} consists of 24 transformer blocks, and each block contains an MLP with two layers with 1024 neurons each. For this study, we consider only the neurons on the output layer. This results in a total of 24,576 MLP neurons whose activations are monitored across the dataset.



We perform Algorithm \ref{alg:feature-selective-neurons} to get highly selective neurons. 
The Shannon Entropy~\cite{shannon1948mathematical} metric helps us identify \textit{interesting} neurons that respond selectively to a small subset of features while remaining inactive for others. Entropy effectively quantifies the degree of bias a neuron exhibits toward particular features. A lower entropy indicates a peaked distribution implying strong preference for a few features while higher entropy suggests a uniform distribution, meaning the neuron responds similarly across diverse features and is therefore less informative for our analysis.


\begin{algorithm}[t]
\caption{Identifying Feature-Selective Neurons in CLIP Vision Encoder}
\label{alg:feature-selective-neurons}
\begin{algorithmic}[1]
\For{each neuron $n$ in all MLP layers}
    \State Rank all images by activation of neuron $n$
    \State Select top-$k$ images with highest activation
    \For{each feature $f_i$ in the dataset}
        \State Count occurrences $o_i$ of $f_i$ across the top-$k$ images
    \EndFor
    \For{each feature $f_i$ in the dataset}
        \State Compute feature affinity value $a_i = \frac{o_i}{\sum_{j=1}^N o_j}$
    \EndFor
    \State Compute Shannon entropy
\EndFor
\State Sort neurons by ascending entropy
\State \Return Neurons with lowest entropy as highly selective neurons
\end{algorithmic}
\end{algorithm}


The Shannon entropy of the feature-occurrence distribution for a neuron is computed as:
\[
\text{Entropy} = \sum_{i = 1}^N -a_i \log a_i
\]
\[
\text{where } a_i = \frac{o_i}{\sum_{j=1}^N o_j}
\]

Here, $a_i$ denotes feature affinity of the neuron to a feature $i$ and it is computed from $o_i$ which denotes the average number of occurrences of feature $i$ among the neuron’s top-$k$ activating images, and $N$ is the total number of features. In our experimental setting, $k = 30$ and $N = 16$, corresponding to 5 shapes, 6 colors, and 5 spatial positions.

Neurons with the lowest entropy values are prioritized for further analysis, as they are more likely to encode disentangled, semantically meaningful features.


%% file: sec/4_experiments.tex

%% file: sec/5_results.tex
\section{Experiments}

\subsection{CLIP on CIFAR}

\begin{figure}
    \centering
    \includegraphics[width=0.9\linewidth]{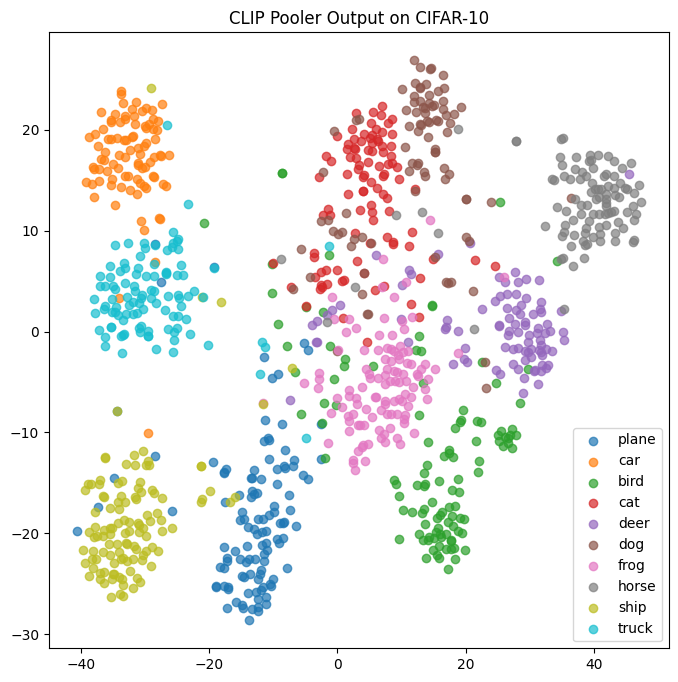}
    \caption{t-SNE plot of CLIP pooler outputs on CIFAR-10. Each point represents a single image embedding colored by its ground-truth class label. }
    \label{fig:cluster-pooler}
\end{figure}

To understand how CLIP’s vision encoder represents image categories, we extract the pooler outputs (final [CLS] embeddings) for a subset of CIFAR-10 images and projected them into 2D space using t-SNE, a non-linear dimensionality reduction technique, to visualize the embedding space. This visualization is shown in Figure \ref{fig:cluster-pooler}.\\

\begin{figure*}[h]
    \centering
    \frame{\includegraphics[width=0.18\linewidth]{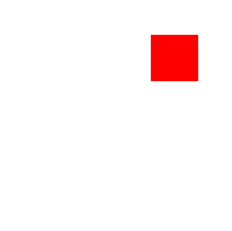}}
    \frame{\includegraphics[width=0.18\linewidth]{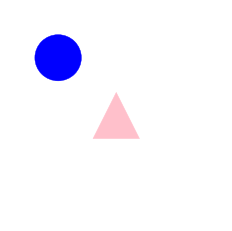}}
    \frame{\includegraphics[width=0.18\linewidth]{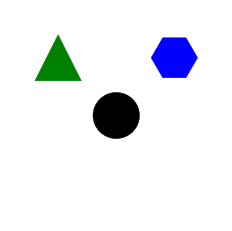}}
    \frame{\includegraphics[width=0.18\linewidth]{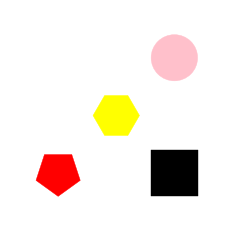}}
    \frame{\includegraphics[width=0.18\linewidth]{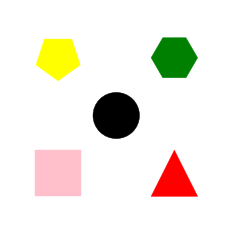}}
    \caption{Sample images from the toy dataset used to study MLP activations in CLIP vision encoder}
    \label{fig:toy-dataset-sample}
\end{figure*}

\begin{table*}[h]
    \centering
    \renewcommand{\arraystretch}{1.2}
    \begin{tabular}{|c|c|c|c|}
        \hline
        \textbf{Original Image} & \textbf{Text Prompt} & \textbf{Grad-CAM Heatmap} & \textbf{Overlayed Heatmap} \\
        \hline
        \multirow{2}{*}{\includegraphics[width=0.22\linewidth]{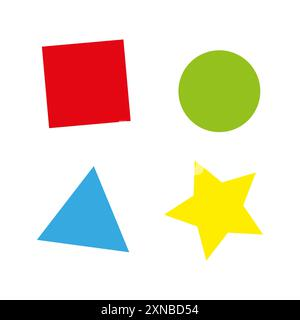}} 
        & \textit{a green circle} 
        & \includegraphics[width=0.22\linewidth]{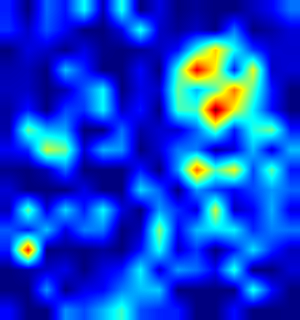} 
        & \includegraphics[width=0.22\linewidth]{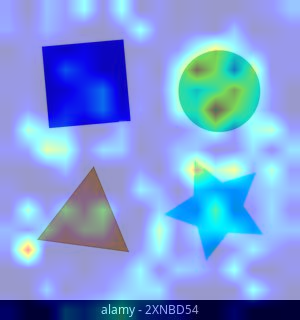} \\
        \cline{2-4}
        & \textit{a green square} 
        & \includegraphics[width=0.22\linewidth]{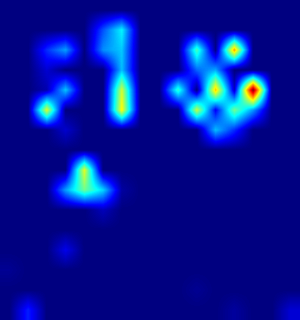} 
        & \includegraphics[width=0.22\linewidth]{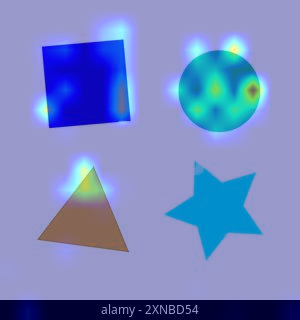} \\
        \hline
        \multirow{2}{*}{\includegraphics[width=0.22\linewidth]{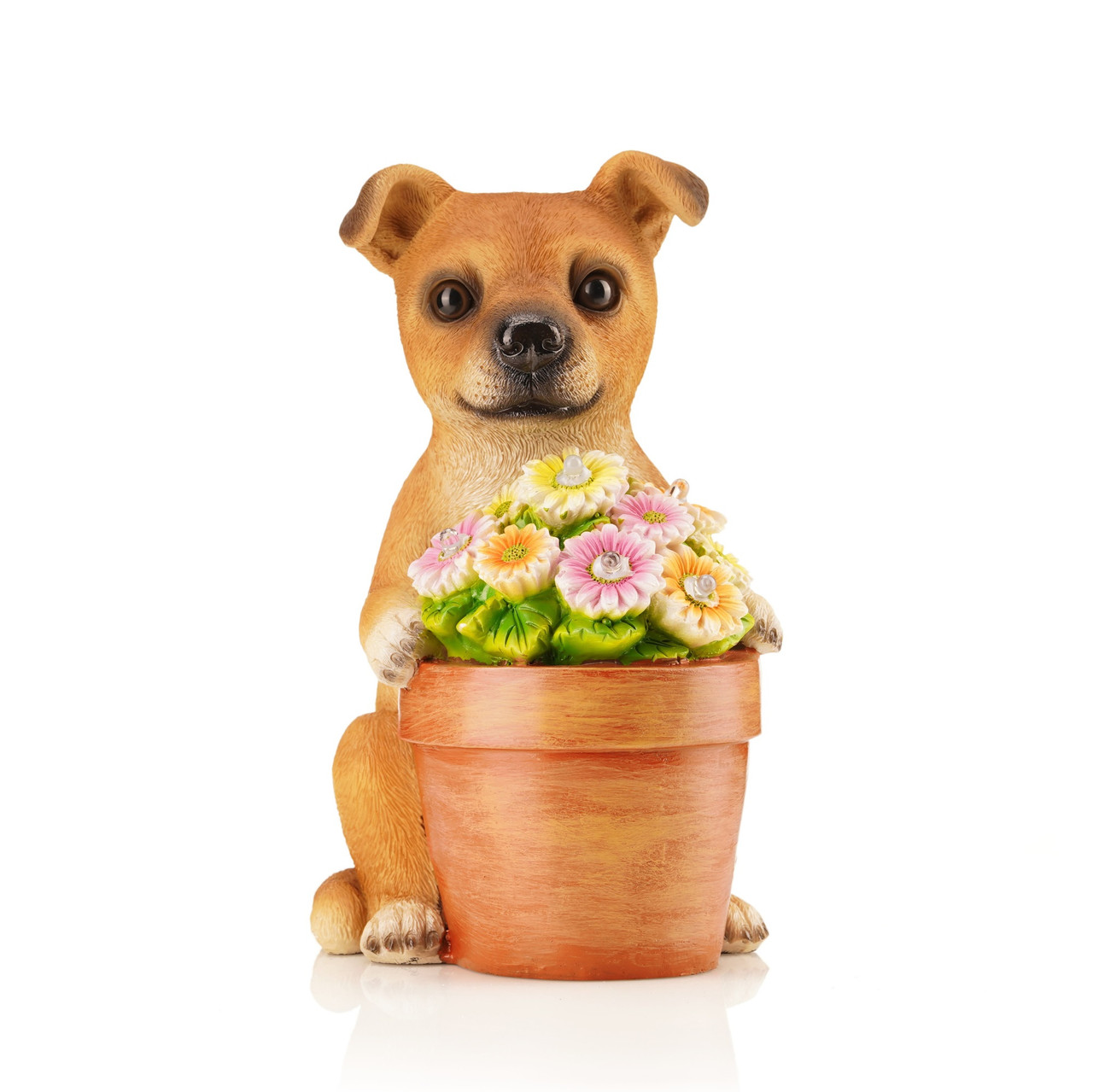}} 
        & \textit{a dog} 
        & \includegraphics[width=0.22\linewidth]{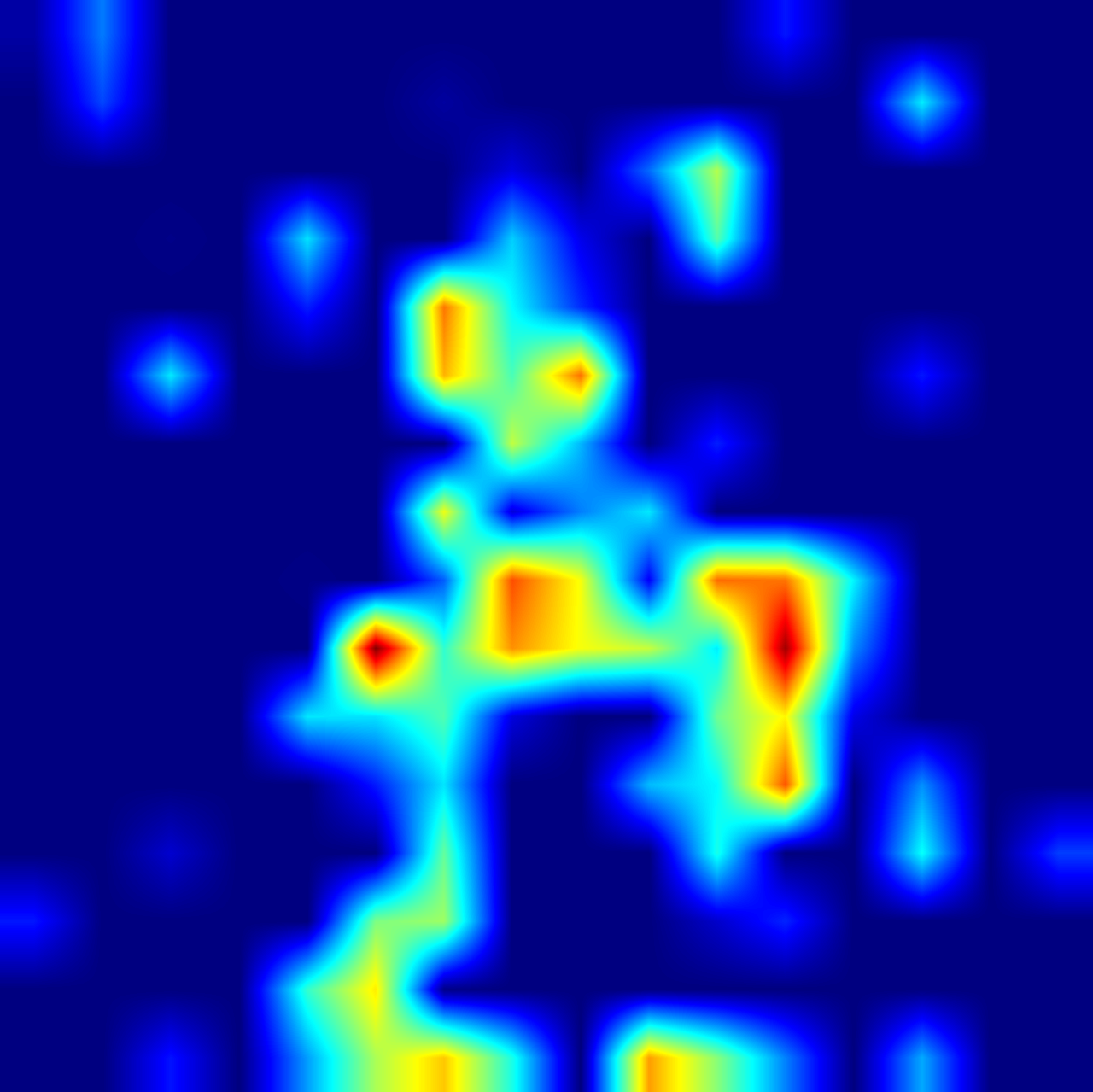} 
        & \includegraphics[width=0.22\linewidth]{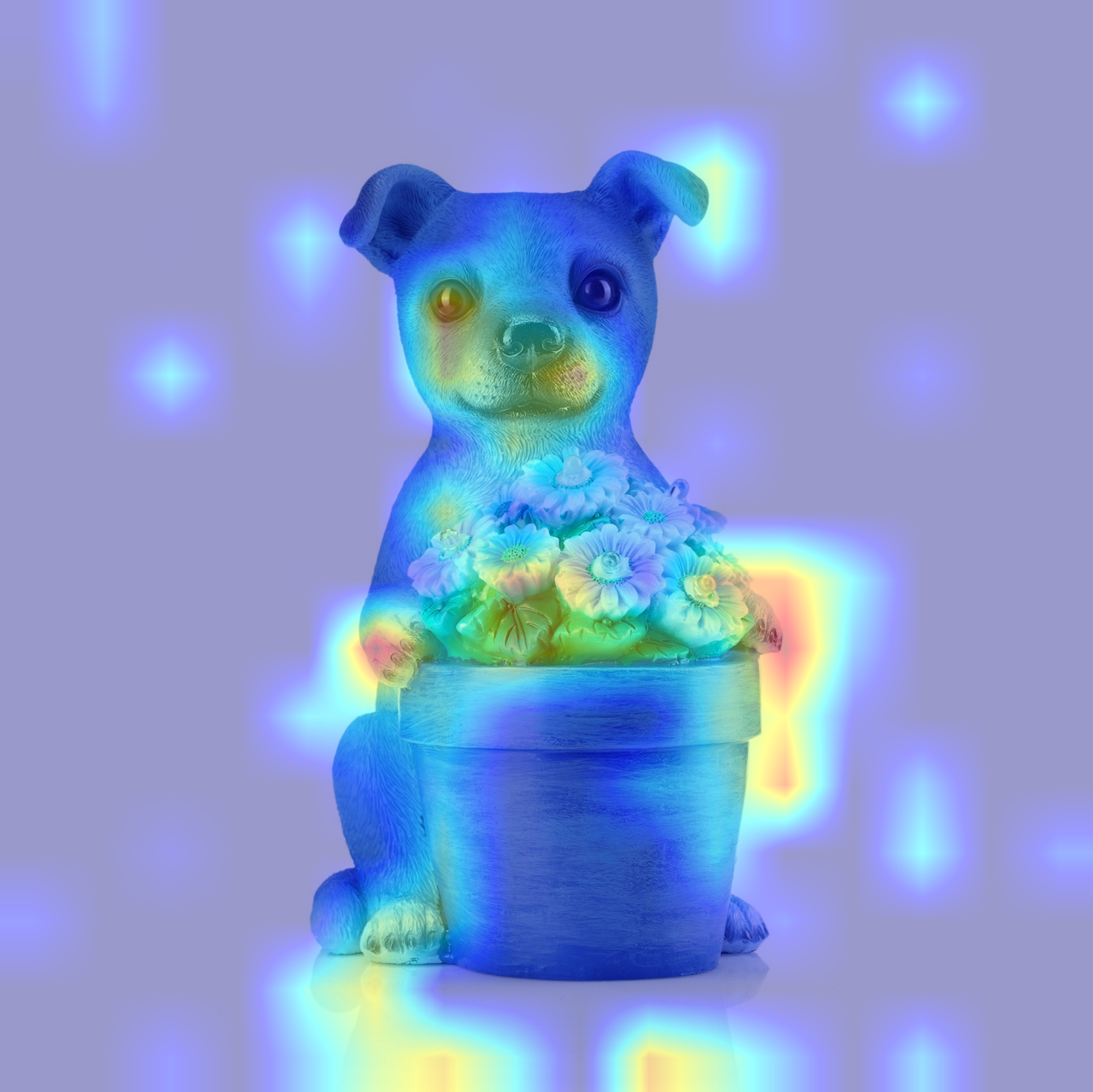} \\
        \cline{2-4}
        & \textit{dog behind pot} 
        & \includegraphics[width=0.22\linewidth]{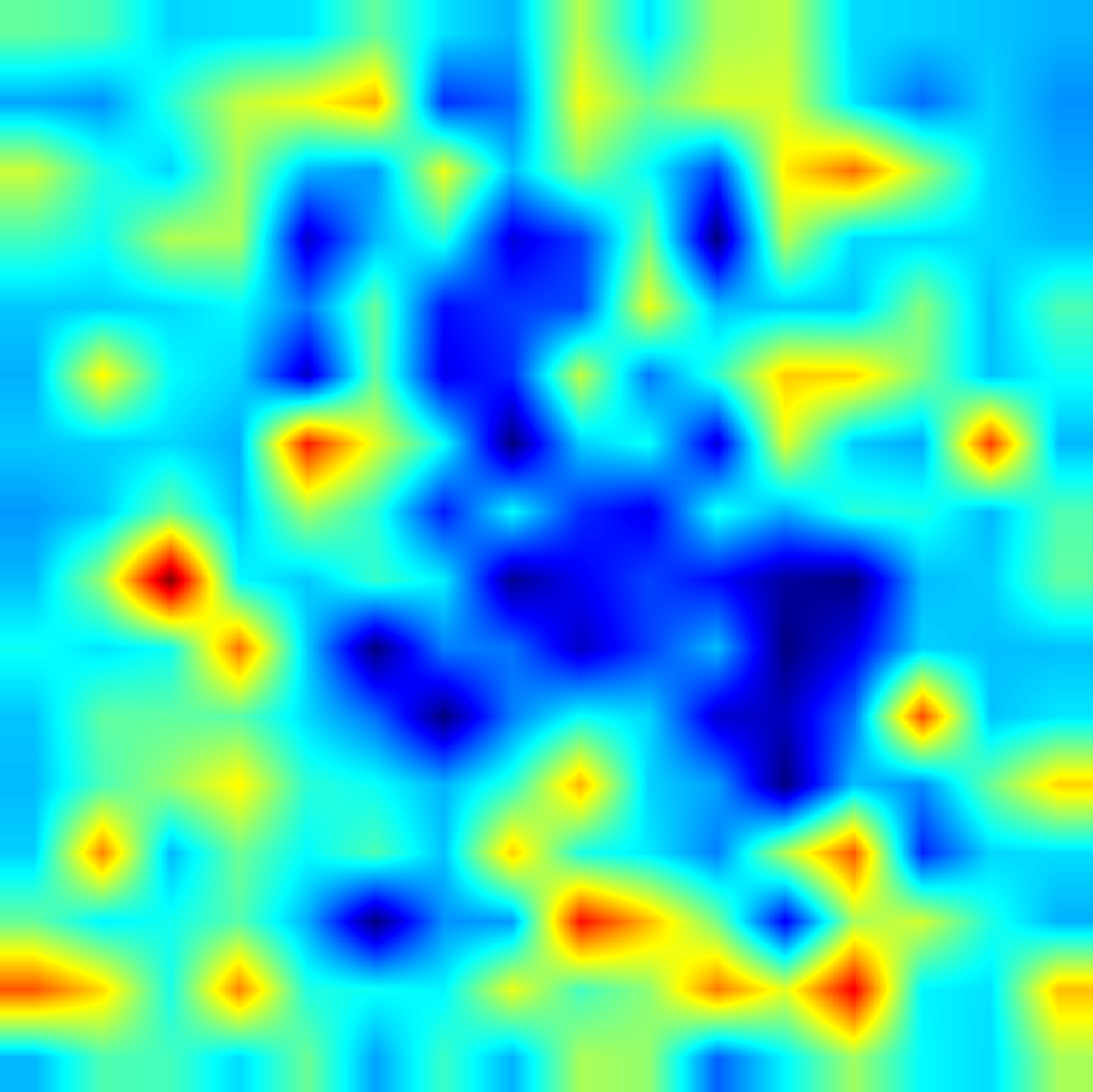} 
        & \includegraphics[width=0.22\linewidth]{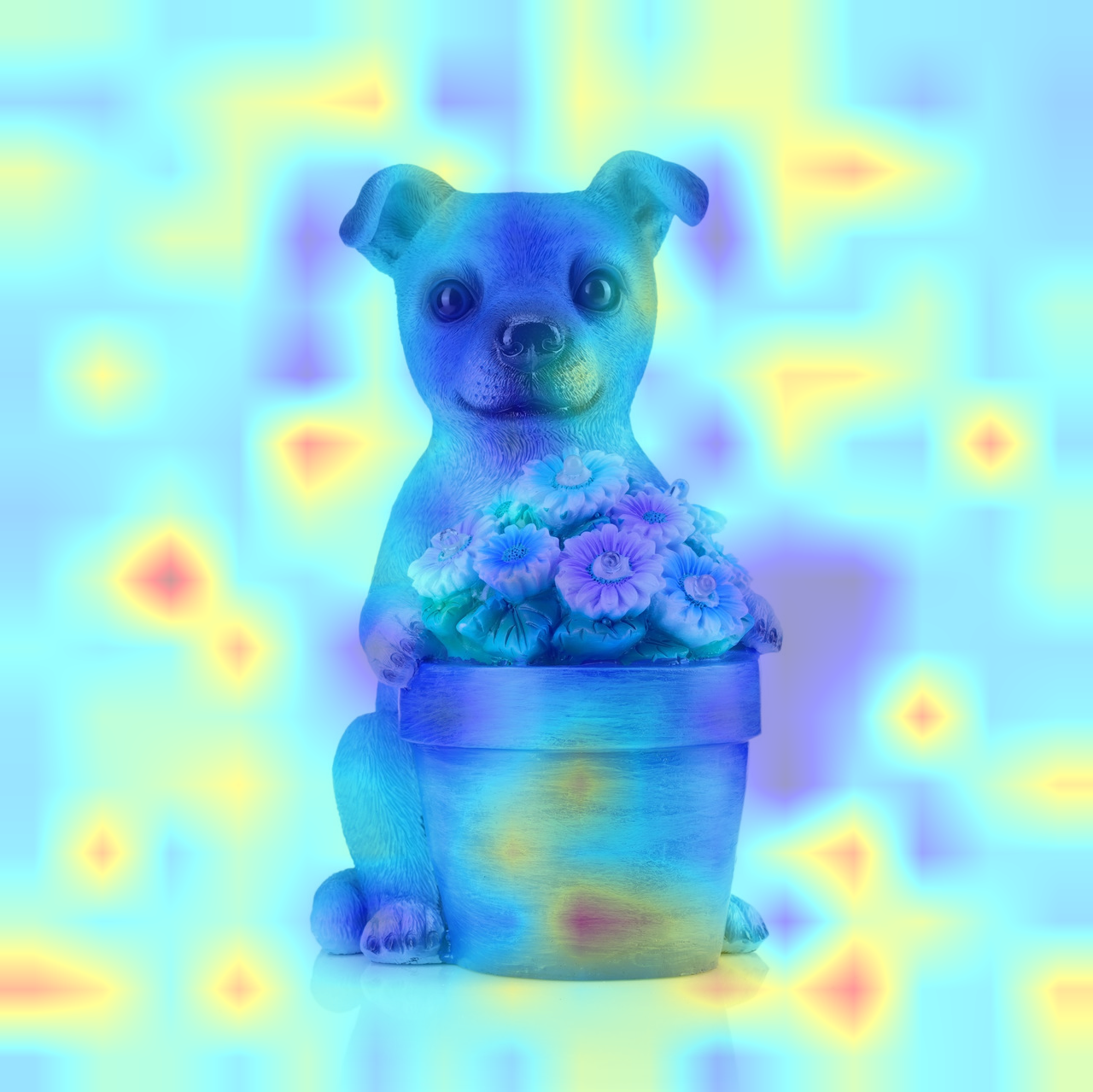} \\
        \hline
    \end{tabular}
    \caption{Grad-CAM results for different text prompts. Each row shows how CLIP's attention shifts for various descriptions of the same image. Incorrect or partial attention localization reveals binding failures (e.g., attending to both green circle and red square for ``green square'' prompt).}
    \label{tab:gradcam_comparisons}
\end{table*}

The figure reveals several key observations:
\begin{enumerate}
    \item Class-specific clusters are clearly separable for most categories, indicating that CLIP’s frozen vision encoder retains discriminative features even when trained without supervision on CIFAR-10.
    \item Some overlap is observed among semantically similar categories (e.g., dog, cat, horse, deer), hinting at potential superposition in the latent space, where related concepts are partially entangled.
\end{enumerate}

These insights validate that the vision encoder itself maintains a strong semantic prior. The proximity between related clusters strongly hints at embedding-level superposition.

We found that because of the rich features of the CIFAR classes, the CLIP neuron activations were harder to analyze. Therefore, we created a simpler dataset, described in \ref{subsec:clip_act_analysis}, consisting of shapes and colors with minimum feature overlap.

\subsection{Analysis of Compositional Failures via Grad-CAM}

To better understand the cause of compositionality failures in CLIP, we begin with a qualitative analysis using Grad-CAM. Grad-CAM is employed here due to its simplicity, transparency and strong grounding in prior interpretability work \cite{compositional_challenges}. Specifically, we ask: when CLIP fails to bind attributes (e.g., color, shape, or spatial relations) to objects, can this be observed directly in its visual attention? This section investigates whether the model's failure modes are reflected in the spatial localization of its attention, as captured by gradient-based attribution.

We evaluate CLIP’s ability to correctly associate multimodal concepts by comparing Grad-CAM visualizations for a set of text prompts applied to the same input image. Table~\ref{tab:gradcam_comparisons} presents these comparisons. Below, we analyze the results case-by-case.

\subsubsection{Geometric Shapes Image}
\textbf{Prompt: \textit{``a green circle''}} \\
This case demonstrates successful grounding: CLIP correctly localizes the green circle, showing that when the prompt exactly matches the image content, the model is capable of precise visual grounding.\\\textbf{Prompt: \textit{``a green square''}} \\
There is no green square in the image; however, CLIP’s attention is split between the red square and the green circle. This illustrates a \textit{binding problem}—the model detects color and shape as separate, unbound features. Rather than suppressing attention when the full conjunction is missing, CLIP incorrectly activates over both partial matches.

\subsubsection{Dog And Pot Image (A Dog Behind a Flower Pot)}
\textbf{Prompt: \textit{``a dog''}} \\
The model successfully identifies the presence of a dog and focuses its attention primarily on the animal. Notably, the Grad-CAM heatmap reveals concentrated activation around semantically salient regions such as the snout, paws, and face. This focused attention on characteristic parts suggests that CLIP’s visual encoder attends to discriminative features when grounding basic object categories. It also indicates that, in the absence of compositional cues or relational modifiers, the model is capable of forming relatively clean object-level representations.\\\textbf{Prompt: \textit{``a dog behind a pot''}} This spatially compositional prompt results in scattered attention across the image, including the dog, the background, and pot regions. CLIP appears unable to interpret and localize relational or positional language in a grounded way, indicating a limitation in \textit{compositional grounding}.

These results demonstrate that CLIP’s internal representations, while effective at coarse image-text alignment, lack robust feature binding. The model tends to process object categories and attributes (e.g., shape, color, spatial relation) independently, often resulting in incorrect or ambiguous localization.


\subsection{MLP Analysis --- Feature Neurons}
\begin{figure}[h]
    \centering
    \includegraphics[width=1\linewidth]{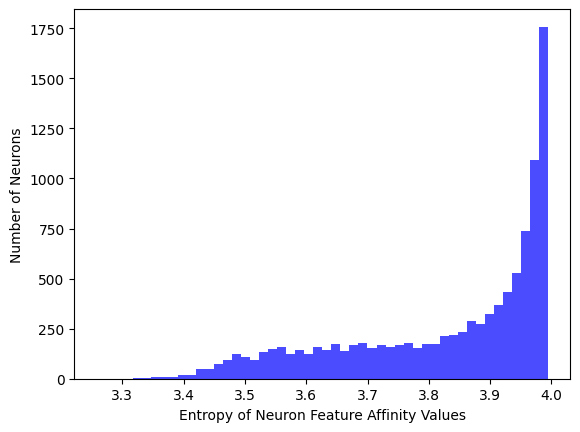}
    \caption{Distribution of the entropy of neuron feature-affinity values in CLIP-ViT-L/14, evaluated on the Toy Shapes dataset.}
    \label{fig:neuron-entropy-hist}
\end{figure}

Through the neuron activation analysis described in Section~\ref{subsec:clip_act_analysis}, we identified several neurons with strong preferences for specific visual features. Figure \ref{fig:neuron-entropy-hist} shows the distribution of the entropy values of all MLP neurons in CLIP-ViT-L/14 computed over the Toy Shapes dataset. A subset of low-entropy neurons—referred to as \textbf{feature neurons}—are presented in Tables \ref{fig:n16563}, \ref{fig:n22447}, \ref{fig:n18466} along with their entropy value \& percentile, top activating features, average activation when the feature is present, and images from the dataset that produce the highest activations in that neuron.

As shown in the tables, the low-entropy neurons exhibit clear selectivity and consistent activation patterns in response to specific features. Their top-activating images predominantly contain a single, distinct object characterized by the target feature, providing strong evidence for their specialization. In contrast, the high-entropy neurons display no clear or consistent visual pattern across their top activations. This does not imply that these neurons are meaningless or uninformative; rather, it is likely that their preferred features are underrepresented or absent in our toy dataset.

\subsection{MLP Analysis --- Evidence of Superposition}

Many of the feature neurons identified in our analysis exhibit signs of \textit{superposition} \cite{elhage2022toy}, that is, they respond to multiple distinct features simultaneously. For example, the $563^{rd}$ neuron in the $16^{th}$ transformer block (highlighted first in Table~\ref{fig:n16563}) shows high activation for both circular and square shapes. To verify this, we analyzed the patch-wise activation maps of this neuron across several test images.

\begin{figure}
\centering
    \includegraphics[width=0.5\textwidth]{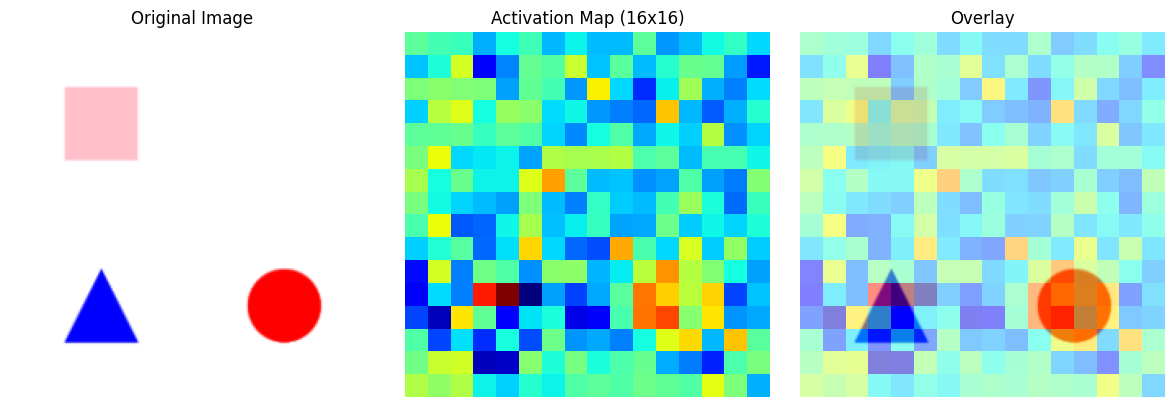}
    \includegraphics[width=0.5\textwidth]{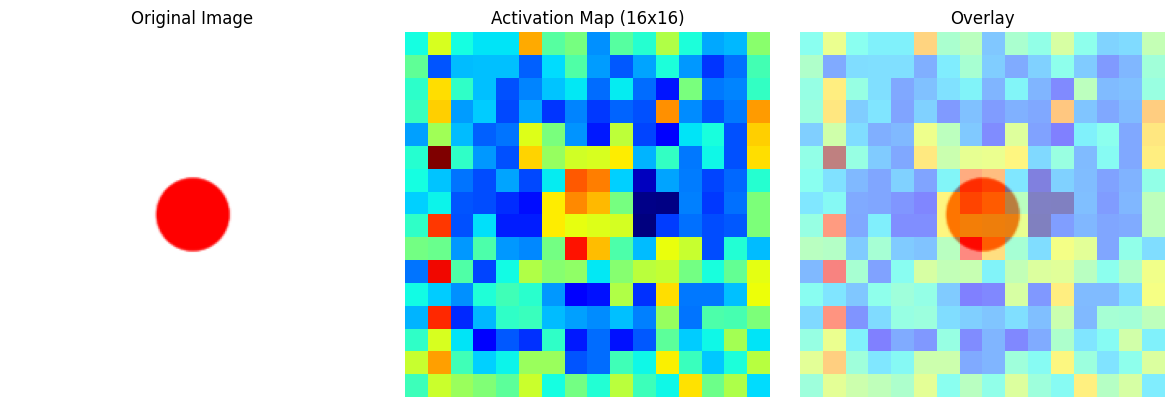}
    \includegraphics[width=0.5\textwidth]{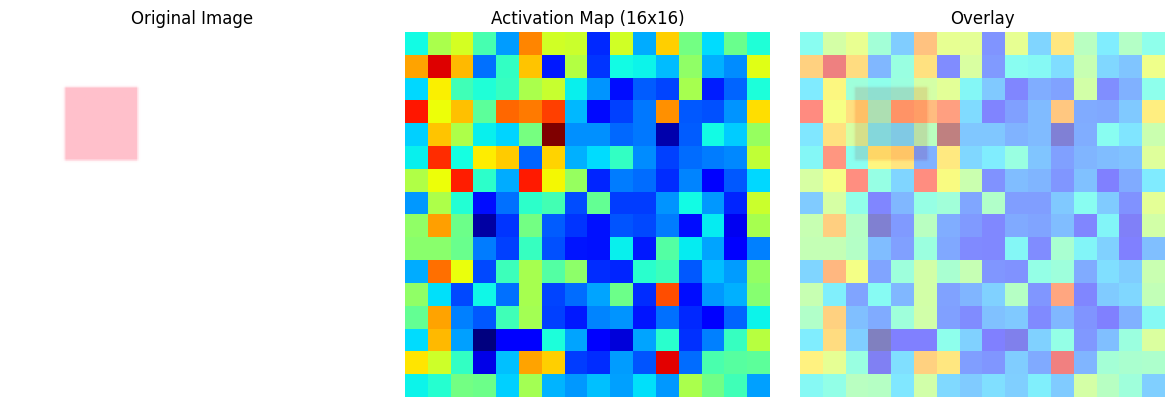}
    \includegraphics[width=0.5\textwidth]{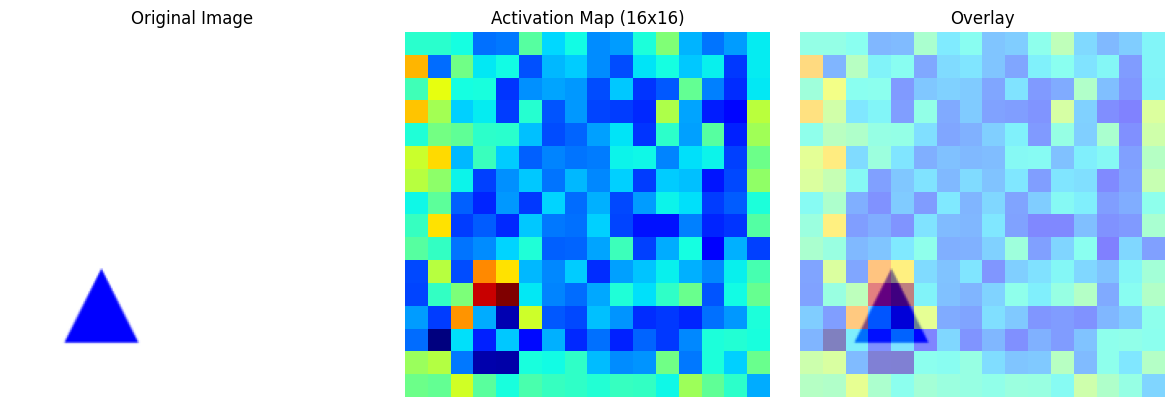}    
    \caption{Images from the toy dataset and the patch-wise activation map of 563rd neuron in layer 16 showing affinity of this neuron to square and circle shapes (first three rows). The neuron shows largely low activations on the bottom-most image containing only a triangle, which is not a target feature for this neuron.}
    \label{fig:patch-wise-activations}
\end{figure}

Figure~\ref{fig:patch-wise-activations} presents a set of images alongside the corresponding activation maps of this neuron. In the first example, we observe that the neuron's activation is strongest (denoted by yellow and red regions) in areas containing its preferred features namely circles, squares, and the colors pink and red. In contrast, regions containing unrelated features such as triangles or blue objects elicit weak responses (blue regions in the heatmap). This suggests the neuron does not indiscriminately fire, but instead responds selectively to a combination of semantically meaningful features.

To further support this observation, we examine activation maps on single-object images. In the second and third rows of Figure~\ref{fig:patch-wise-activations}, the neuron activates strongly only in the presence of one or more of its preferred features. In the final row, where the image contains none of the relevant features, the activation is low throughout the entire spatial map indicated by uniformly blue regions.

These results provide strong evidence that the neuron simultaneously encodes multiple visual concepts, validating the presence of superposition. Understanding such behavior is crucial for downstream compositional reasoning, as neurons that entangle multiple features may introduce ambiguity when precise attribute–object bindings are required.

As a side note, one reason for the activations in Figure \ref{fig:patch-wise-activations} are not very spatially focused on the locations of the objects in the image, is that image tokens pass through many layers of self-attention, which mixes and redistributes information across tokens. As a result, neurons in the later layers may respond to features that are no longer in the same location as they were in the original image. Another possible reason, one that complements the first, is that ViTs often use background tokens as places to gather and process information. Darcet et al.\cite{registers} found that ViTs tend to select less important tokens, often from background regions, and use them as pooling spots to combine information from other tokens.


\begin{table*}
\centering
    \begin{tabularx}{\textwidth}{| c |}
        \hline
        \textbf{16th Layer 563rd Neuron \textbar\ Entropy: 3.36 (0.18 \%ile)}\\ \hline
        \textbf{Top Features: } Circle (0.57), Pink (0.57), Square (0.53), Top-Left (0.4), Red (0.3)\\
        \hline
            \begin{minipage}{0.975\textwidth}
                \centering
                \includegraphics[width=\textwidth]{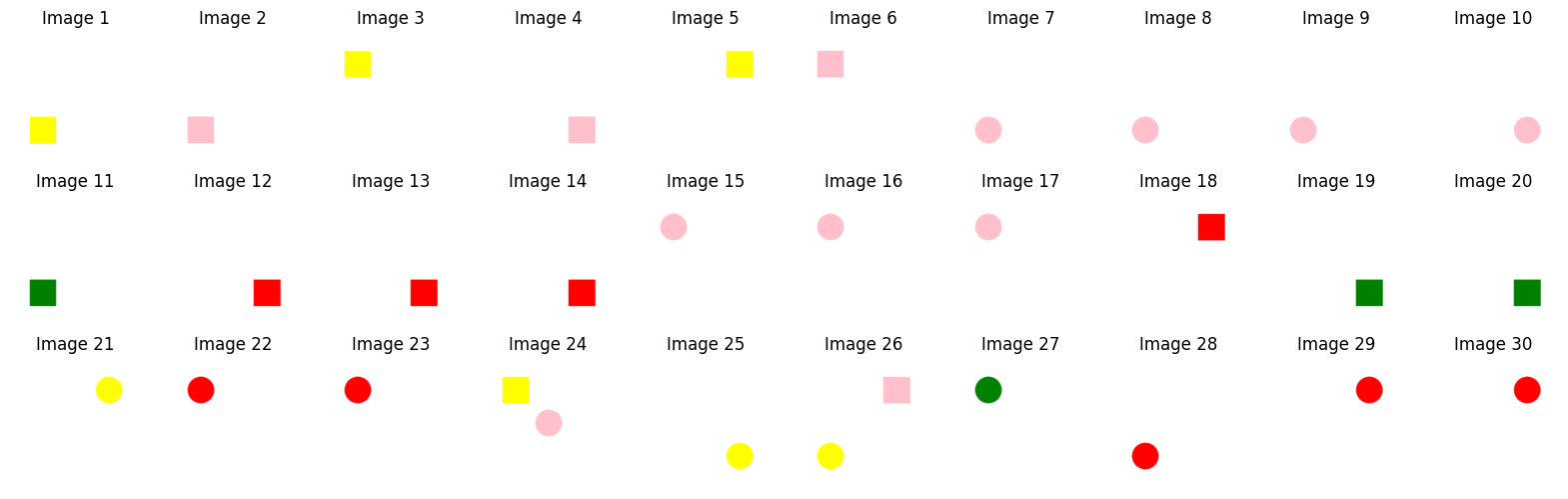}
            \end{minipage}\\ 
        \hline
    \end{tabularx}
    \caption{Top features and Top activating images for 563rd Neuron in the 16th Layer. This neuron activates the most when handling images containing square and circle shapes and the color pink.}
    \label{fig:n16563}
\end{table*}

\begin{table*}
\centering
    \begin{tabularx}{\textwidth}{| c |}
        \hline
        \textbf{22nd Layer 447th Neuron \textbar\ Entropy: 3.34 (0.08 \%ile)}\\ \hline
        \textbf{Top Features: } Circle (0.6), Middle (0.6), Pentagon (0.47), Pink (0.37), Top-Left (0.3)\\\hline
            \begin{minipage}{0.975\textwidth}
                \centering
                \includegraphics[width=\textwidth]{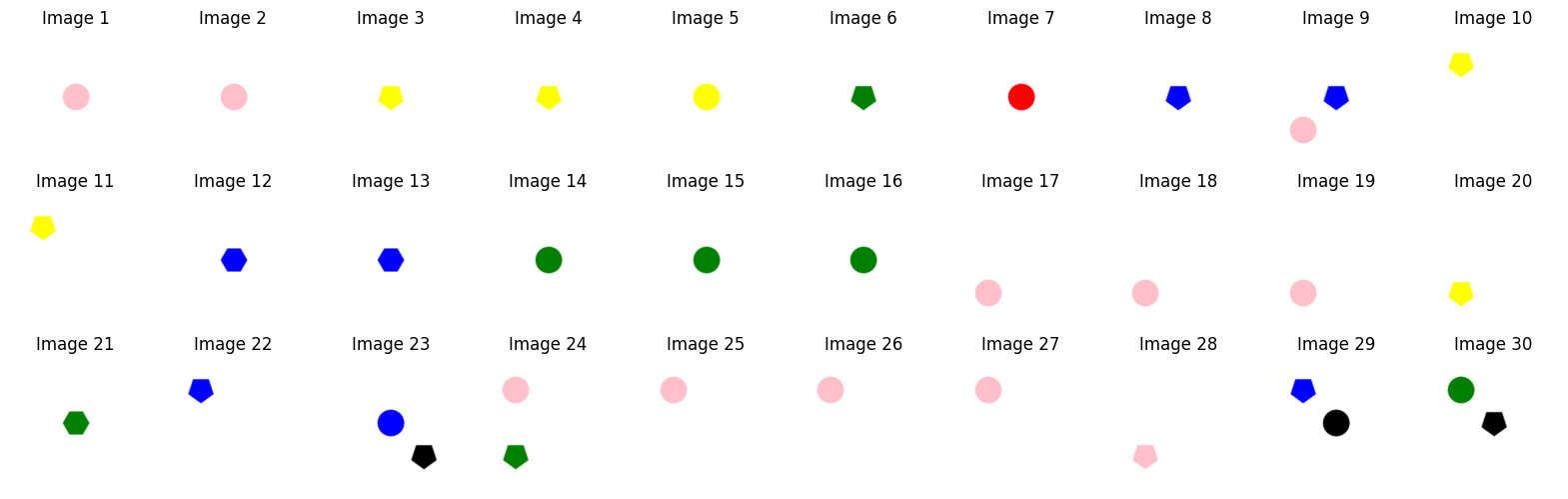}
            \end{minipage}\\ 
        \hline
    \end{tabularx}
    \caption{Top features and Top activating images for 447th Neuron in the 22nd Layer. This neuron activates the most when handling images containing the circle and pentagon shapes and the middle position.}
    \label{fig:n22447}
\end{table*}

\begin{table*}
\centering
    \begin{tabularx}{\textwidth}{| c |}
        \hline
        \textbf{18th Layer 466th Neuron \textbar\ Entropy: 3.4 (0.38 \%ile)}\\ \hline
        \textbf{Top Features: } Pink (0.63), Black (0.4), Square (0.37), Bottom-Right (0.37), Pentagon (0.3)\\\hline
        \begin{minipage}{0.975\textwidth}
                \centering
                \includegraphics[width=\textwidth]{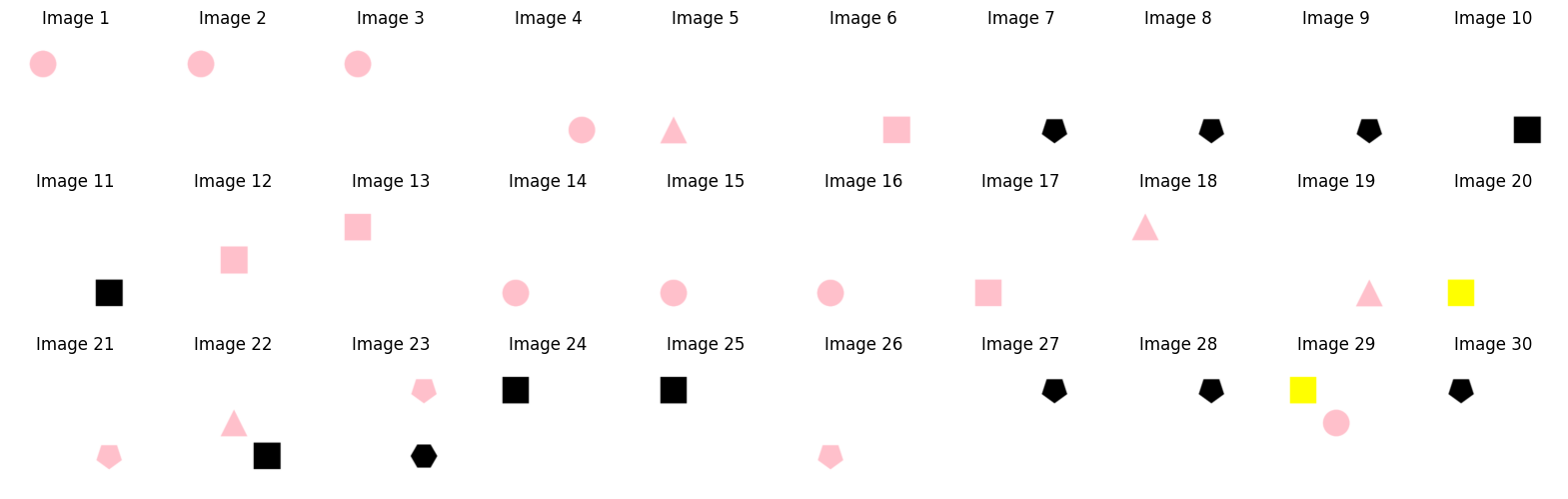}
            \end{minipage}\\ 
        \hline
    \end{tabularx}
    \caption{Top features and Top activating images for 466th Neuron in the 18th Layer. This neuron activates the most when handling images containing pink and black shapes.}
    \label{fig:n18466}
\end{table*}

\subsection{Effects of Superposition}

\subsubsection{Motivation}
Building on the evidence of superposition, we hypothesize that such neuron-level superposition erodes the model’s ability to form clean, compositional object–attribute bindings. In other words, if the same low-entropy neuron fires for both “red” and “square”, the model may confuse a red circle for a green square.

\subsubsection{Method}
Taken from the 16 features in our shapes dataset, we create pairs of features. For every ordered feature pair
\(
\langle f_{1},f_{2}\rangle
\) we:

\begin{enumerate}[leftmargin=1.5em]
  \item \textbf{Quantify superposition} using for the 1000 lowest-entropy neurons. The measure for Superposition or \textbf{S}, measured between features $f_1$ and $f_2$ is 
              \[
               \textbf{S}(f_1, f_2) = \sum_{i=1}^{n}
                 \frac{a_{f_1}^{(i)} + a_{f_2}^{(i)}}
                      {\sum_{j=1}^{16}{a_{f_j}^{(i)}}}
            \]
    where $a_{f_j}^{(i)}$ is the affinity of feature $j$ to neuron $i$.
            
  \item \textbf{Quantify compositional separability} of the corresponding image embeddings. The metrics used for this are 
        \begin{itemize}[leftmargin=1.2em]
          \item Cluster-center distance \textbf{D}\((f_{1},f_{2})\): the Euclidean distance between the two single-feature centroids (larger is better).
          \item Misclassification rate \textbf{M}\((f_{1},f_{2})\): fraction of embeddings whose nearest centroid is of the \emph{wrong} feature (smaller is better).
        \end{itemize}        
\end{enumerate}



\subsubsection{Results}
\begin{figure}[h]
    \centering
    \includegraphics[width=1\linewidth]{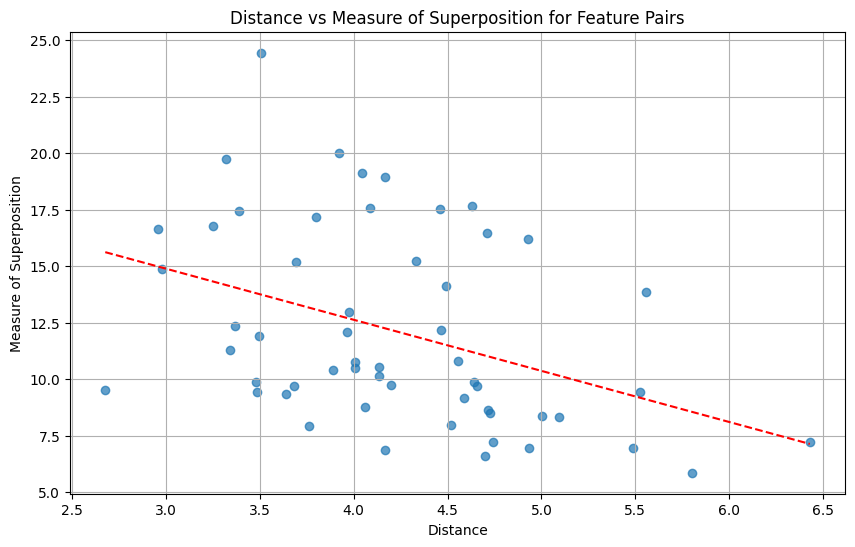}
    \caption{Distance (\textbf{D}) vs Measure of Superposition (\textbf{S}) for Feature Pairs}
    \label{fig:superpos-1}
\end{figure}

\begin{figure}[h]
    \centering
    \includegraphics[width=1\linewidth]{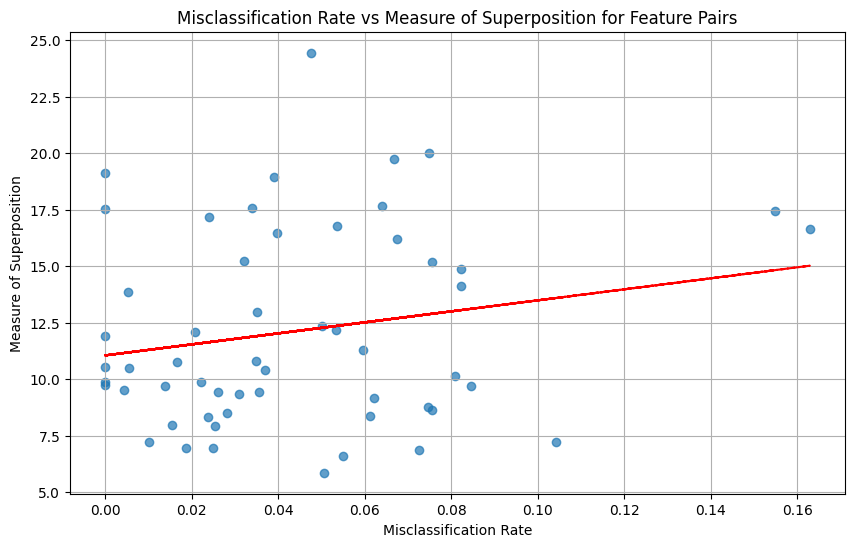}
    \caption{Misclassification Rate (\textbf{M}) vs Measure of Superposition (\textbf{S}) for Feature Pairs}
    \label{fig:superpos-2}
\end{figure}

Scatter plots of Measure of Superposition or $S$ versus $D$ and $M$ are shown in Figures \ref{fig:superpos-1}, \ref{fig:superpos-2} respectively.

The key observations from these plots are -
\begin{enumerate}
    \item \emph{Inverse relation with distance.}  
        Feature pairs that are superimposed more (high $S$) exhibit \textbf{smaller cluster gaps} (low $D$) in embedding space.
        This indicates that superposition pulls concept representations closer together, reducing their geometric separability.
     \item \emph{Direct relation with error.}  
        The same high-$S$ pairs suffer \textbf{higher misclassification rates} $M$, showing that entangled neurons translate into increased attribute–object binding mistakes.
\end{enumerate}

The experiment demonstrates a connection, albeit a weak correlation that warrants further investigation, between superposition inside CLIP’s MLP layers and its external difficulties with compositional reasoning.

%% file: sec/6_conclusion.tex
\section{Limitations and Future Work}
This study is limited to the CLIP-ViT-L/14 model, whereas most modern vision-language models (VLMs) adopt variants such as SigLIP\cite{smolvlm} and alternative architectures like SAM \cite{deepseekvl}. Future work could strengthen our claims by establishing a more direct causal link between weak compositional feature representations and failures in compositional reasoning. Additionally, extending our analysis to more complex, feature-annotated datasets would facilitate a deeper investigation into the relationship between neuronal feature superposition and compositional reasoning limitations.

\section{Conclusion}
Our study uncovers a mechanistic connection between internal feature representation and CLIP’s image embeddings.

\begin{enumerate}[leftmargin=1.5em,itemsep=0.4em]
  \item \textbf{Neuron-level superposition.}  
        Feature entanglement is present not only at the embedding-level but
        all the way down to individual neurons, many of which may encode multiple, semantically unrelated attributes.
  \item \textbf{Impact on compositionality.}  
        The stronger this superposition, the weaker CLIP’s ability to bind objects and attributes: high entanglement predicts smaller embedding-space separation and higher misclassification rates on compositional tasks.
\end{enumerate}

These findings establish superposition as a key bottleneck for object–attribute compositionality and motivate future work on disentangling neuron activations to improve CLIP’s feature representation.